\documentclass{article} % For LaTeX2e
\usepackage{nips14,times}
\usepackage{hyperref}
\usepackage{url}
\usepackage{amssymb}
\usepackage{amsfonts}
\usepackage{eulervm}
\usepackage{multirow}
\usepackage{graphicx}
  
\title{Feature Weight Tuning for Recursive Neural Networks}
\usepackage{times}
\usepackage{latexsym}
\usepackage{url}
\usepackage{color}
\usepackage{multirow}
\usepackage{comment}
\usepackage{amsmath} 

\usepackage{amssymb}

\usepackage{etoolbox}
\usepackage{graphicx}
 
 \usepackage{enumitem}

\author{Jiwei Li \\
Computer Science Department\\
Stanford University\\
Stanford, CA, USA 94305\\
jiweil@stanford.edu}

\nipsfinalcopy  
\begin{document}

\maketitle

\begin{abstract}
This paper addresses how a recursive neural network model can automatically leave out useless information and emphasize important evidence, in other words, to perform ``weight tuning" 
for higher-level representation acquisition.
We propose two models, Weighted Neural Network (WNN) and Binary-Expectation Neural Network (BENN), which automatically control how much one specific unit contributes to the higher-level representation. 
The proposed model can be viewed as incorporating a more powerful compositional function for embedding acquisition in recursive neural networks.
Experimental results demonstrate the significant improvement over standard neural models.
\end{abstract}

\section{Introduction}
\begin{comment}
When people compare Support Vector Machine \cite{cortes1995support} (or SVM based approaches, e.g., kernels, regularizations) with deep learning architectures, one distinct difference is that
the former first identifies a fixed set of artificial features from feature selection process, 
and 
then trains the associated weight values from an optimization framework.
While the latter, 
automatically learns the features, which are characterized in an abstract, low-dimensioned and usually human-uninterpretable vectors. 
These vectors, along with the parameters in the architecture, 
are trained through optimizing the task-specific objective function.
As
the task-specific objective function leads to task-specific features, the learned features are sometimes more powerful 
in capturing what is really needed in the correspondent NLP tasks \cite{hinton2013features}. 
\end{comment}

\begin{figure} 
\centering
\includegraphics[width=3in]{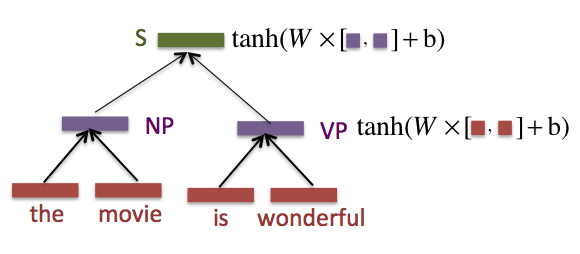}
\caption{Illustration of Standard Recursive Neural Network for Sentence-level Representation Calculation. }
\label{fig1}
\end{figure}

Recursive neural network models \cite{williams1989learning} constitute one type of neural structure for obtaining higher-level representations beyond word-level such as phrases or sentences. It works in a bottom-up fashion on tree structures (e.g., parse trees) in which long-term dependency can be to some extent captured.
Figure \ref{fig1} gives a brief illustration about how recursive neural models work to obtain the distributed representation for the short sentence ``{\it The movie is \underline{wonderful}}''. 
Suppose $h_{\text{is}}$ and $h_{\text{wonderful}}$ are the embeddings for tokens {\it is} and {\it wonderful}. The representation for their parent node VP at second layer is given by:
\begin{equation}
h_{VP}=f (W\cdot [h_{is}, h_{wonderful}]+b)
\label{equ1}
\end{equation}
where $W$ and $b$ denote parameters involved in the convolutional function. $f(\cdot)$ is the activation function, usually $tanh$ or $sigmod$ or the rectifier linear function.

For NLP tasks, the obtained embeddings could be further fed into 
task-specific
machine learning models\footnote{Of course, embeddings could also be optimized through the task-specific objective functions.}, through which parameters are to be optimized. 
Take sentiment analysis as an example, 
we could feed the aforementioned sentence embedding into a logistic regression model to classify it as either positive or negative.
Embeddings are sometimes more capable of capturing latent semantic meanings or syntactic rules within the text than manually developed features, from which many NLP tasks would benefit (e.g., \cite{socher2013recursive,irsoy2013bidirectional}).

\begin{comment}
For many deep learning (DL) applications in NLP tasks,
neural architectures usually require the aforementioned type of vector representations to represent the units.
The unit could be
tokens/words  (e.g., \cite{bengio2006neural,collobert2008unified,mikolov2013linguistic}), N-grams \cite{socher2011semi}, phrases \cite{socher2013recursive}, sentences (e.g., \cite{irsoy2013bidirectional,blunsom2014convolutional,limodel1}), discourse \cite{jirepresentation,lirecursive}, paragraphs \cite{le2014distributed} or documents \cite{denil2014modelling}.
\end{comment}

\begin{table}
\centering
\begin{tabular}{|c|c|}\hline
Model&Accuracy\\\hline
unigram SVM & 0.743\\\hline
Recursive Neural Net&0.730\\\hline
\end{tabular}
\caption{A brief comparison between SVM and standard neural network models for sentence-level sentiment classification using date set from \cite{pang2002thumbs}. Neural network
models are trained with L2 regularization, using AdaGrad \cite{duchi2011adaptive} with minibatches
(for details about implementations of recursive networks, please see Section 2). 
Parameters are trained based on 5-fold cross validation on the training data.
We report the best performance searching optimum regularization parameter, optimum batch size for mini-batches and convolutional function.
Word embeddings are borrowed from Glove \cite{jeffreypenningtonglove} with dimensionality of 300, which generates better performance than word2vect, SENNA \cite{collobert2011natural} and RNNLM \cite{mikolov2011rnnlm}. 
}
\label{tab1}
\end{table}

Such a type of structure suffers some sorts of intrinsic drawbacks. 
Revisit Figure \ref{fig1}, common sense tells us that  
tokens like ``{\it the}'', ``{\it movie}'' and ``{\it is}"  do not contribute much to the sentiment decision but
word ``{\it wonderful}" is the key part 
(and a good machine learning model
should have the ability of learning these rules). 
Unfortunately, the intrinsic structure of recursive neural networks makes it less flexible to get rid of the influence from less sentiment-related tokens.
If the keyword ``wonderful" hides too deep in the parse tree, 
for example, as in the sentence ``{\it I studied Russia in Moscow, where all my family think the winter is \underline{wonderful}}",
it will takes quite a few convolution steps before the keyword `wonderful" comes up to the surface, with the consequence that its influence on the final sentence representation could be very trivial.
Such an issue, usually referred to as gradient vanishing \cite{bengio1994learning}.  
 is not specific for recursive models, but for most deep learning architectures.

When we compare neural models with SVM,  
one notable weakness of big-of-word based SVM is its inability of considering how words are combined to form meanings (or order information in other words) \cite{mooney2005subsequence}.
But interestingly, such downside of SVM comes with the advantage of resilience in feature managing as the optimization is ``flat-expanded". 
Low weights will be assigned to  less-informative evidence,
which could further be 
 pushed to zero by regularization. 
Table \ref{tab1} gives a brief comparison between unigram based SVM and neural network models for sentence-level sentiment prediction on Pang et al.'s dataset \cite{pang2002thumbs}, and as can be seen, in this task, standard neural network models underperform SVM\footnote{To note, results here are not comparable with Socher et al.'s work \cite{socher2013recursive} which obtains state-of-art performance in sentiment classification, as here labels at sentence-level constitute only sort of supervision for both SVM and neural network models (for details, see footnote 7).}.

Revisit the form of Equ.\ref{equ1}, 
there are two straws we can grasp at to deal with the aforementioned problem: (1) expecting
the learned feature embeddings for less useful words such as $the$\footnote{We just use this example for illustration. Practically, $the$ might be a good sentiment
indicator as it usually co-appears with superlatives.
} exert very little influence (for example, a zero vector for the best) (2) expecting the compositional parameters $W$ and $b$ are extremely powerful.
For the former, it is sometimes hard, as mostly we borrow (or initialize) word embeddings from those trained from large corpus (e.g.,
 word2vec, RNNLM \cite{mikolov2011rnnlm,mikolov2010recurrent}, SENNA \cite{collobert2011natural}),  
rather than training embeddings from task-specific objective functions as neural models can be easily over fitted given the small amount of training data\footnote{There are cases, for example, \cite{socher2013recursive}, where task-specific word embeddings are learned. But it requires sufficient training data to avoid over fitting. 
For example, Socher et al.'s work labels every single node as positive/negative/neutral along parse trees (with a total number of more than 200,000 phrases).}. 

Regarding the latter issue, several alternative compositional functions have been proposed to enable more varieties in composition to
cater.
Recent proposed approaches include, for example, Matrix-Vector RNN \cite{socher2012semantic}, which represents every word as
both a vector and a matrix, RNTN \cite{socher2013recursive} which allows greater interactions between the input
vectors, and the algorithm presented in \cite{lirecursive} which associates different labels (e.g., POS tags, relation tags) with different sets of compositional parameters.
These approaches to some extent enlarge the power of compositional functions.

In this paper, we borrow the idea of ``weight tuning" from feature based SVM and try to incorporate such idea into neural architectures.
To achieve this goal, we propose two recursive neural architectures, Weighted Neural Network (WNN) and Binary-Expectation Neural Network (BENN). 
The major idea involved in the proposed approaches is to associate each node in the recursive network with additional parameters,
indicating how important  it is for final decision. 
For example, we would expect such type of a structure would dilute the influence of tokens like ``the" and ``movie" but magnifies the impact of tokens like ``wonderful" and ``great" in sentiment analysis tasks.
Parameters associated with proposed models are automatically optimized through the objective function manifested by the data.
The proposed model combines the capability of neural models to capture the local compositional meanings with weight tuning approach to
reduce the influence of undesirable information 
at the same time, and yield better performances 
 in a range of different NLP tasks when compared with standard neural models.

The rest of this paper is organized as follows: Section 2 briefly describes the related work. The details of WNN and BENN 
are illustrated in Section 4 and experimental results are presented in Section 5, followed by a brief conclusion.

\section{Related Work}
Distributed representations, calculated based on neural frameworks, are extended beyond 
token-level, to represent N-grams \cite{socher2011semi}, phrases \cite{socher2013recursive}, sentences (e.g., \cite{irsoy2013bidirectional,blunsom2014convolutional}), discourse \cite{jirepresentation,lirecursive}, paragraphs \cite{le2014distributed} or documents \cite{denil2014modelling}.
Recursive and recurrent \cite{schuster1997bidirectional,sutskever2011generating} models constitute two types of commonly used frameworks for sentence-level embedding acquisition. 
Different variations of recurrent/recursive models are proposed to cater for different scenarios (e.g., \cite{irsoy2013bidirectional,socher2013recursive}).
Other recently proposed approaches included sentence compositional approach proposed  in \cite{KalchbrennerACL2014}, or paragraph/sentence vector \cite{le2014distributed} where representations are optimized through predicting words within the sentence. 

Neural network architecture sometimes requires a vector representation of each input token.   Various deep learning architectures have been explored to learn these embeddings in an unsupervised manner from a large corpus \cite{bengio2006neural,collobert2008unified,mnih2007three,mikolov2013efficient}, which might have different generalization capabilities and are able to capture the semantic meanings depending on the specific task at hand.  

Both of the proposed architectures are in this work inspired by the long short-term memory (LSTM) model, first proposed by Hochreiter and Schmidhuber back in 1990s \cite{hochreiter1997long,gers2000learning} to process time sequence data 
where there are very long time lags of unknown size between important events\footnote{\url{http://en.wikipedia.org/wiki/Long_short_term_memory}}.
LSTM associates each time with a series of ``gates" to determine
whether the information from early time-sequence should be forgotten \cite{gers2000learning}
and when current information should be allowed to flow into or out of the memory.
LSTM could partially address gradient vanishing problem in recurrent neural models and have been widely used in machine translation \cite{sundermeyer2014translation,cho2014learning}

\section{``Weight Tuning" for Neural Network}
Let $s$ denote a sequence of token $s=\{w_1,w_2,...,w_{n_s}\}$. 
It could be phrases, sentences etc. 
Each word $w$ is associated with a specific vector embedding ${\bf e_w}=\{e_w^1,e_w^2,...,e_w^K\}$, where $K$ denotes the dimensionality of the word embedding.   We wish to compute the vector representation for sentence $s$, denoted as
 $h_s=\{h_s^1,h_s^2,...,h_s^K\}$ based on parse trees using recursive neural models.
Parse tree for each sentence is obtained from Stanford Parser \cite{socher2013parsing}.

\subsection{WNN for Recursive Neural Network}
For any node $C$ in the parse tree, it is associated with representation $h_C$.
The basic idea of WNN is to associate each node $C$ with an additional weight variable $M_C$, which is in range (0,1), to denote the importance of current node.
Technically, $M_C$ is used to pushing the output representation of not-useful node towards the direction of 0 and retain relatively important information.

We expect that information regarding the importance of current node (e.g., whether it is relevant to positive/negative sentiment)
is embedded in its representation $h_C$. So we use a convolution function to enable this type of information to emerge to the surface from the following compositional functions:

\begin{equation}
R_C =f (W_{M}\cdot h_C+b_{M})
\end{equation}
\begin{equation}
M_C=sigmod (U_M^T\cdot R_C) 
\end{equation}
where $W_M$ is a $D\times K$ dimensional matrix and $b_B$ is the $1\times K$ bias vector.
$R_C$ is a $K$ dimensional intermediate vector.
Such implementation can be viewed as using a three-layer neural model with $D$ latent neurons for an output
projected to a [0,1] space.

Let $output(C)$ denote the output from node $C$ to its parent. 
In WNN, $output(C)$ would consider both current information, which is embedded in the embedding $h_C$ and its related importance $M_C$.  $output(C)$ is therefore given by
\begin{equation}
output(C)=M_C*h_C
\end{equation}
Recall the example in Figure \ref{fig1}, we have:
\begin{equation}
\begin{aligned}
&output(the)=M_{the}\cdot h_{the}\\
&output(movie)=M_{movie}\cdot h_{movie}
\end{aligned}
\end{equation}
If the model thinks not too much relevant information embedded in $h_C$, the value of $M_C$ would be small, pushing the output vector towards 0. 
 The representations for parents, for example $VP$ and $NP$ in Figure \ref{fig1}, are therefore computed as follows:
\begin{equation}
\begin{aligned}
&h_{VP}=tanh(W_B\cdot [output(is), output(wonderful)])\\
&h_{NP}=tanh(W_B\cdot [output(the), output(movie)])
\end{aligned}
\label{equ7}
\end{equation}
where $W_B$ denotes a $K\times 2K$ dimensional matrix and
 $[output(is), output(wonderful)]$ denotes the concatenation of the two vectors. 
In an optimum situation, 
 $M_{the}$ and $M_{movie}$ will take the values around 0, leading to the representation of node $NP$ to an around-zero vector.

\paragraph{Training WNN}  
For illustration purpose, we use a binary classification task to show how to train WNN. 
To note, the described training approach
applies to other situations (e.g., multi-class classification, regression) with minor adjustments. 

In a binary classification task,
each sequence is associated with a gold-standard label $y_s$. $y_s$ takes value of 1 if positive and 0 otherwise. 
Standardly, to determine the value of $y_s$, 
we feed the representation $h_s$ into a logistic regression model:
\begin{equation}
p(y_s=1)=sigmod (U^T h_s+b)
\label{equ8}
\end{equation}
where $U^T$ is a $1*K$ vector and $b$ denotes the bias. 
Then
by adding the regularization part parameterized by $Q$, the loss function $J(\Theta)$ for the training dataset is given by:
\begin{equation}
J(\Theta)=- \log [p(y_s=1)^{y_s}\cdot (1-p(y_s=1))^{1-y_s}]\
+Q\sum_{\theta\in\Theta}\theta^2
\end{equation}
Revisit the example in Figure \ref{fig1}, 
for any parameter $\theta$ to optimize, the calculation for gradient $\partial J/\partial\theta$ 
is trivial once $\partial [M_{VP}\cdot h_{VP}]/\partial\theta$ and $\partial [M_{NP}\cdot h_{NP}]/\partial\theta$ are obtained, which are given by:
\begin{equation}
\frac{\partial M_{VP}\cdot h_{VP}}{\partial\theta}=M_{VP}\frac{\partial h_{VP}}{\partial\theta}+\frac{\partial M_{VP}}{\partial\theta}h_{VP}
\label{equ10}
\end{equation}
To note, $h_{VP}$ is embraced in $M_{VP}$.
As all components in Equ \ref{equ10} are continuous, the gradient can be efficiently obtained
from standard backpropagation \cite{goller1996learning,socher2010learning}.

\begin{figure*} 
\centering
\includegraphics[width=5in]{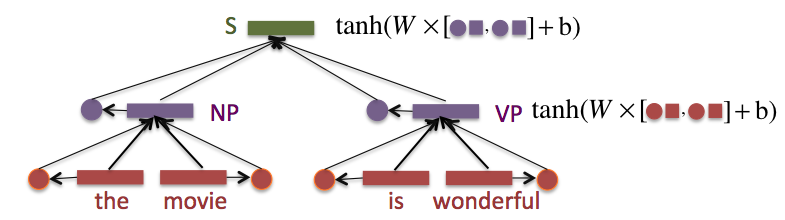}
\caption{Illustration of WNN. }
\end{figure*}

\subsection{BENN for Recursive Neural Network}
BENN  
associates each node with a binary variable $B_C$, which is sampled from a binary distribution parameterized by $L_C$.
$L_C$ is a scalar fixed to the range of [0,1], indicating the possibility that current node should pass information to its ancestors.
$L_C$ is obtained in the similar ways as in WNN by using a convolution to project the current representation $h_C$ to a scalar lying within [0,1].
\begin{equation}
R_C =f (W_{B}\cdot h_C+b_{B})
\end{equation}
\begin{equation}
L_C=sigmod (U_B^T\cdot R_C) 
\end{equation}
\begin{equation}
B_C\sim binary(L_C)
\end{equation}
For smoothing purpose,
in BENN, current node $C$ outputs the expectation of embedding $h_C$ to its parent, as given by:
\begin{equation}
output(C)=E[h_C]
\end{equation}
Take the case in Figure \ref{fig1} as an example again, vector $h_{NP}$ will therefore follow the following distribution:
\begin{equation}
\begin{aligned}
&p(h_{NP}=tanh(W_B[h_{the},h_{movie}]))=L_{the}\cdot L_{movie}\\
&p(h_{NP}=tanh(W_B[0,h_{movie}]))=(1-L_{the})\cdot L_{movie}\\
&p(h_{NP}=tanh(W_B[h_{the},0]))=L_{the}\cdot (1-L_{movie})\\
&p(h_{NP}=tanh(W_B[0,0]))=(1-L_{the})\cdot (1-L_{movie})\\
\end{aligned}
\end{equation}
$E[h_{NP}]$ can be further obtained based on such distribution 
\begin{equation}
E[h_{NP}]=\sum_{h}P(h_{NP}=h) \cdot h
\end{equation}
To note, for leaf nodes, $E[h_C]=h_C$.

\paragraph{Training BENN}
For training, we again use binary sentiment classification for illustration.
For any sentence $s$ with label $y_s$, we have
\begin{equation}
p(y_s=1)=sigmod (U^T E[h_s]+b)
 \end{equation}
With respect to any given parameter $\theta$, the derivative of $E[h_s]$ is further given by:
\begin{equation}
\begin{aligned}
\frac{\partial E(h_s)}{\partial\theta}
&=\frac{\partial L_{NP}\cdot L_{VP}\cdot tanh(W_B[E(h_{NP}),E(h_{VP})])}{\partial\theta}\\
&+\frac{\partial (1-L_{NP})\cdot L_{VP}\cdot tanh(W_B[0,E(h_{VP})])}{\partial\theta}\\
&+\frac{\partial L_{NP}\cdot (1-L_{VP})\cdot tanh(W_B[E(h_{NP}),0]}{\partial\theta}\\
&+\frac{\partial (1-L_{NP})\cdot (1-L_{VP})\cdot tanh(W_B[0,0])}{\partial\theta}
\end{aligned}
\end{equation}
With all components being continuous, the gradient can be efficiently obtained
from standard backpropagation.

\begin{comment}
\subsection{BENN for Recurrent Neural Network}
For recurrent models, given current word embedding $e_i$, similarly, we have the current output:
\begin{equation}
\begin{aligned}
&output(h_{i})=E[h_i]\\
&=tanh(W\cdot E[h_{h_{i-1}}]+V\cdot e_i+b)\cdot L_{i-1}\cdot L_{e_i}\\
&=tanh(W\cdot E[h_{h_{i-1}}]+V\cdot 0+b)\cdot L_{i-1}\cdot (1-L_{e_i})\\
&=tanh(W\cdot 0+V\cdot e_i+b)\cdot (1-L_{i-1})\cdot L_{e_i}\\
&=tanh(W\cdot 0+V\cdot 0+b)\cdot (1-L_{i-1})\cdot (1-L_{e_i})\\
\end{aligned}
\end{equation}
where we have
\begin{equation}
\begin{aligned}
&L_{i-1}=sigmod(U_B^T f(W_B\cdot E[h_{i-1}])+b_B)\\
&L_{e_i}=sigmod(U_B^T f(W_B\cdot e_i)+b_B)
\end{aligned}
\end{equation}
\end{comment}

\section{Experiment}
We perform experiments to better understand the behavior of the proposed models compared with standard neural models (and other variations).
To achieve this, we implement our model on problems that
require fixed-length vector representations for phrases or sentences.

\subsection{Sentiment Analysis}
\paragraph{Sentence-level Labels}
We first perform experiments on 
dateset from \cite{pang2002thumbs}. In this setting, binary labels at the top of sentence constitute the only resource of supervision (to note, it is different from settings described in \cite{socher2013recursive}).
All neural models adopt the same settings for fair comparison: L2 regularization, gradient decent based on AdaGrad with mini batch size of 25, tuned parameters for regularization on 5-fold cross validation. 

For standard neural models, we implement two settings: standard (GLOVE) where word embeddings are directly fixed to GLOVE and standard (learned) where word embeddings are treated as parameters to optimize in the framework. 
Additionally, we implemented some recent popular variations of recursive models with more sophisticatedly designed compositional functions, including: 
\begin{itemize}
\item MV-RNN (Matrix-Vector RNN): which was proposed in \cite{socher2012semantic}
which
represents every node in a parse tree as both a vector and
a matrix. Given the vector representation $h_{C_1}$, matrix representation $V_{C_1}$ for child node $C_1$,  $h_{C_2}$ and $V_{C_2}$
for child node $C_2$, the vector representation $h_{p}$ and matrix representation $V_{p}$ for parent $p$ are given by:
\begin{equation}
\begin{aligned}
&h_{p}=f(W_1[V_{C_1}\cdot h_{C_2},V_{C_2}\cdot h_{C_1}])\\
&V_{p}=f(W_1[V_{C_1},V_{C_2}])
\end{aligned} 
\end{equation}
We fix word vector embeddings using SENNA and treat matrix representations as parameters to optimize. 
\item  RNTN (Recursive Neural Tensor Network): proposed in \cite{socher2013recursive}. Given $h_{C_1}$ and $h_{C_2}$ for children nodes, RNTN computes parent vector $h_p$ in the following way:
\begin{equation}
h_p=f([h_{C_1}, h_{C_2}]^T V [h_{C_1}, h_{C_2}]+ W [h_{C_1}, h_{C_2}])
\end{equation}
\item Label-specific: associate
each of the sentence roles (i.e., VP, NP or NN) with a specific composition matrix.
\end{itemize}

\begin{table}
\centering
\begin{tabular}{|c|c|c|}
\hline
\multicolumn{2}{|c|}{Model}           & Accuracy \\ \hline
\multirow{6}{*}{Recursive} & Standard (GLOVE) & 0.730    \\ \cline{2-3} 
			  & Standard (learned) & 0.658    \\ \cline{2-3} 
                           & MV-RNN   & 0.704    \\ \cline{2-3} 
                           & RNTN     & 0.760    \\ \cline{2-3} 
                           & Label-Specific&0.768 \\ \cline{2-3} 
                           & WNN      & 0.778    \\ \cline{2-3} 
                           & BENN      &0.772    \\\hline
SVM                        & unigram  & 0.743    \\ \hline
\end{tabular}
\caption{Binary Sentiment Classification with Supervision only at Sentence Level. Word embeddings are initialized from 300 dimensional embeddings borrowed from GLOVE \cite{jeffreypenningtonglove}.}
\label{tab2}
\end{table}

\begin{figure*} 
\centering
\includegraphics[width=5in]{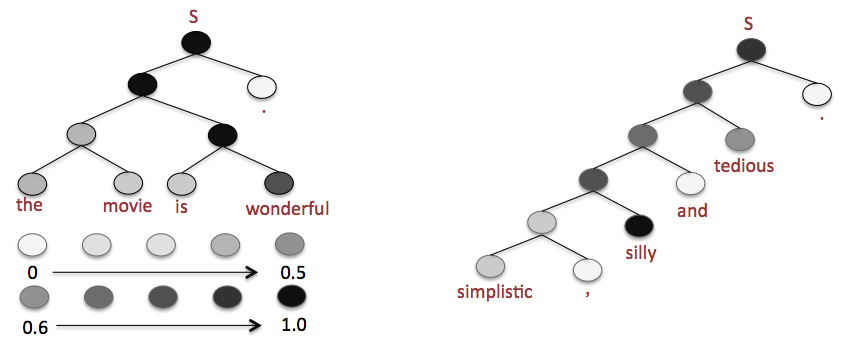}
\caption{Visual illustration of automatic learning of weight $M_C$ associated with each node in WMM. }
\label{fig3}
\end{figure*}

We report results in Table \ref{tab2}. As discussed earlier, 
standard neural models
underperform the bag of word models.
To note, for derivations of standard neural models such as Standard (learned) and MV-RNN with many more parameters to learn, 
 the performance is 
 even worse due to over-fitting. 
WNN and BENN
, although not significantly output bag of words SVM, generates better results, yielding significant improvement over standard neural models and existing revised versions. 
Figure \ref{fig3} illustrates the automatic learned muted factor $M_C$ regarding different nodes in the parse tree based on recursive network. As we can observe , the model is capable of learning the proper weight of vocabularies, assigning larger  weight values to important sentiment indicators (e.g., wonderful, silly and tedious) and suppressing the influence of less important ones. We attribute the better performance of proposed models over standard neural models to such automatic weight-tuning ability. 

To note, in this scenario, we are not claiming that we generate state-of-art results using the proposed model. More sophisticated bag-of-word models, for example, (e.g., \cite{maas2011learning}) can generate better performance that what the proposed models achieve. The point we wish to illustrate here is that the proposed 
models provide a promising perspective 
 over standard neural models due to the ``weight tuning" property. And in the cases where more detailed data is available to capture the compositionally, the proposed models hold promise to generate more compelling results, as we will illustrate in Socher et al's setting for sentiment analysis.

\paragraph{Socher et al's settings} We now consider Socher et al's dataset \cite{socher2013recursive} for sentiment analysis, where contains gold-standard labels at every phrase node in the parse tree.
The task could be considered either as a 5-way fine-grained classification task where
the labels are very-negative/negative/neutral/positive/very-positive or a 2-way coarse-way as positive/negative based on labeled dataset. 
We follow the experimental
protocols described in \cite{socher2013recursive} (word embeddings are treated as parameters to learn rather than fixed to externally borrowed embeddings).
In this
work we only consider labeling the full sentences.

In addition to varieties of neural models mentioned in Socher et al's work, we also report the performance of recently proposed {\it paragraph vector model} \cite{le2014distributed}, 
which first obtains sentence embeddings in an unsupervised manner by predicting words within the context and then feeds the pre-obtained embeddings into
a logistic regression model. 
{\it paragraph vector} achieves current the state-of-art performance regarding Socher et al's dataset.

Performances are reported in Table \ref{tab3}. As can be seen,
the proposed approach slightly underperforms current state-of-art performance achieved by {\it paragraph vector} 
but outperforms
all the other versions of recursive neural models, indicating the 
adding ``weight tuning" parameters indeed leads to better compositionally. 

To note, when there is more comprehensive dataset which we can rely on to obtain the favorable task-specific word embeddings, 
compositionally plays an important role 
in deciding whether the review is positive or negative
by harnessing local word order information.
In that case, neural models exhibit its power in capturing local evidence from the composition,
leading to significantly better performance 
than all
bag-of-words based models (i.e., SVM and Bigram Naives Bayes).

\begin{table}
\small
\centering
\begin{tabular}{|c|c|c|}\hline
Model&Fine-grained&Coarse-grained\\\hline
SVM& 0.407&0.794\\\hline
Bigram Naives&0.419&0.831\\\hline
Recursive&0.432&0.824\\\hline
MV-RNN&0.444&0.829\\\hline
RNTN&0.457&0.854\\\hline
Paragraph Vector&0.487&0.878\\\hline
WNN&0.482&0.865\\\hline
BENN&0.475&0.870\\\hline
\end{tabular}
\caption{The performance of proposed approaches compared with other methods on Stanford sentiment treebank dataset. Baseline performances are reported from \cite{socher2013recursive,le2014distributed}.}
\label{tab3}
\end{table}

\subsection{Document-level Sentiment Analysis on IMDB dataset}
We move on to sentiment analysis at document level. We use the IMDB dataset proposed by Maas
et al. \cite{maas2011learning}. The dataset consists of 100,000 movie reviews taken
from IMDB and each movie review contains several sentences.
We follow the experimental protocols described in \cite{maas2011learning}.

We first train word vectors from word2vect 
using the 75,000 training documents. Next we train the compositional functions using the 25,000 labeled documents by keeping the word embedding fixed.
We first obtain sentence-level representations using WNN/BENN (recursive).
As each review contains multiple sentences, we convolute sentence representations to one single vector using WNN/BENN recurrent network. 
We cross validate parameters using the labeled documents and test the models on the 25,000 testing reviews.

The results of our approach and other baselines are reported in Table \ref{tab5}.
 As can be seen, for long documents, bag-of-words (both unigram and diagram) perform
quite well and it is difficult to beat.
Standard neural models again do not generate competent results compared with bag of word models in this task. But by incorporating weighted tuning mechanism, we got much better performance, roughly $5\%$ when compared against standard neural models. Although WNN and BENN still underperform current state-of-art model Paragraph Vector \cite{le2014distributed}, they produces better performance than bag-of-word models. 

\begin{table}
\centering
\begin{tabular}{|c|c|}\hline
Model&Precision\\\hline
SVM-unigram&0.869\\\hline
SVM-bigram&0.892\\\hline
recursive+recurrent&0.870\\\hline
WNN&0.902\\\hline
BENN&0.910\\\hline
\end{tabular}
\caption{The performance of proposed model compared to other approaches on binary classification on IMDB dataset. Results for baselines are reported from \cite{maas2011learning}. To note, the reported results here underperform current state-of-the-art performances. Paragraph vectors \cite{le2014distributed} reported an accuracy of 92.58 in terms of IMDB dataset. }
\label{tab5}
\end{table}

\subsection{Sentence Representations for Coherence Evaluation}
Sentiment analysis forces more on the semantic perspective of meaning. Next we turn to a more syntactic oriented task, 
where we obtain sentence-level representations based on the proposed model to decide the coherence of a given sequence of sentences.

We use corpora widely employed for coherence prediction \cite{barzilay2004catching,barzilay2008modeling}.  
One contains reports on airplane accidents from the National Transportation Safety Board and the other contains reports about earthquakes from the Associated Press. Standardly, we use pairs of articles, one containing the original
document order which is assumed to be coherent and used as positive examples, and the other a random permutation
of the sentences from the same document, which are treated as not-coherent examples.
We follow the protocols introduced in \cite{barzilay2004catching,louis2012coherence,limodel} by considering a window approach and feeding the concatenation of representations of adjacent sentences into a logistic regression model, to be classified as either coherent or non-coherent. 
In test time, we assume that the model makes a right decision if the original
document gets a score higher than the one with random permutations. Current state-of-art performance regarding this task is obtained by using standard recursive network as described in \cite{limodel}.

Table \ref{tab5} illustrates the performance of different models. Entity-grid model \cite{barzilay2004catching} generates state-of-art performance among all non-neural network models. As can be seen, neural models perform pretty well in this task when compared against existing feature based algorithm.
From the reported results, better sentence representations are obtained by incorporating ``weighted tuning" properties, pushing the state of art of this task to the accuracy of 0.936. 

\begin{table}

\centering
\begin{tabular}{|c|c|}\hline
Model& Accuracy\\\hline
WNN-recursive&0.930\\\hline
BENN-recursive&0.936\\\hline
recursive&0.920\\\hline
Entity-Grid&0.888\\\hline
\end{tabular}
\caption{Comparison of Different Coherence models. Reported baseline results are reprinted from \cite{barzilay2004catching}.}
\label{tab5}
\end{table}

\section{Conclusion}
In this paper, we propose two revised versions of neural models, WNN and BENN for obtaining higher level feature representations for a sequence of tokens.
The proposed framework automatically incorporates the concept of ``weight tuning" of SVM into the DL architectures which lead to better higher-level representations and generate better performance against standard neural models in multiple tasks.
While it still underperforms bag-of-word models in some cases, and the newly proposed paragraph vector approach,  it provides as an 
alternative to  
existing recursive neural models for 
representation learning.

To note, while we limit our attentions to recursive models in this work, the idea of weight tuning in WNN and BENN, that associates
nodes in neural models with additional 
 weighed variables is a general one and can be extended to many other deep learning models with minor adjustment. 
 
\bibliographystyle{unsrt}
\bibliography{acl}

\end{document}